\documentclass[11pt]{article}
\usepackage{fullpage}
\usepackage{amsmath,amssymb}
\usepackage{wrapfig}
\usepackage{algorithm}
\usepackage[noend]{algorithmic}
\usepackage{amsmath}
\usepackage{amssymb}
\usepackage{amsthm}
\usepackage{bbm}
\usepackage{bm}
\usepackage{color}
\usepackage{dsfont}
\usepackage{enumerate}
\usepackage{graphicx}
\usepackage{subfigure}
\usepackage{times}
\usepackage[usenames,dvipsnames]{xcolor}

\usepackage[capitalize,noabbrev]{cleveref}

\usepackage[utf8]{inputenc} 
\usepackage[T1]{fontenc}    
\usepackage{url}            
\usepackage{booktabs}       
\usepackage{amsfonts}       
\usepackage{nicefrac}       
\usepackage{microtype}      
\usepackage{amsopn}
\usepackage{bbm}
\usepackage{enumerate}

\newtheorem{theorem}{Theorem}

\newtheorem{lemma}{Lemma}

\newcommand{\PP}{\mathbb{P}}
\newcommand{\EE}[1]{\mathbb{E}\left[#1\right]}
\newcommand{\Prob}[1]{\mathbb{P}\left(#1\right)}
\newcommand{\E}{\mathbb{E}}
\newcommand{\one}[1]{\mathbb{I}\left\{#1\right\}}
\DeclareMathOperator{\sgn}{sgn} 
\DeclareMathOperator{\argmin}{argmin}

\newcommand{\R}{\mathbb{R}}
\newcommand{\N}{\mathbb{N}}
\newcommand{\pms}{\{\pm 1\}}
\newcommand{\oR}{\overline{R}}

\newcommand{\Rad}{\text{Rad}}

\newcommand{\defeq}{\doteq}

\title{Crowdsourcing with Sparsely Interacting Workers
\thanks{Authors appear in alphabetical order. This material is based upon work supported in part by  Division of Systems Engineering Post-doctoral Fellowship, Boston University, and by Saligrama's NSF Grants CCF: 1320566, CNS: 1330008, CCF: 1527618, the U.S. Department of Homeland Security, Science and Technology Directorate, Office of University Programs, under Grant Award 2013-ST-061-ED0001, by ONR contract N00014-13-C-0288 and NGA-NURI Grant HM1582-09-1-0037. The views and conclusions contained in this document are those of the authors and should not be interpreted as necessarily representing the social policies, either expressed or implied by our sponsors.}}

\author{Yao Ma, Alex Olshevsky, Venkatesh Saligrama, Csaba Szepesvari}

\begin{document}

\maketitle

\begin{abstract}
We consider estimation of worker skills from worker-task interaction data (with unknown labels) for the single-coin crowd-sourcing binary classification model in symmetric noise. 
We define the (worker) interaction graph whose nodes are workers and an edge between two nodes indicates 
whether or not the two workers participated in a common task. 
We show that skills are asymptotically identifiable if and only if an appropriate limiting version of the interaction graph is irreducible and has odd-cycles. 
We then formulate a weighted rank-one optimization problem to estimate skills based on observations on an irreducible, aperiodic interaction graph. 
We propose a gradient descent scheme and show that for such interaction graphs estimates converge asymptotically 
to the global minimum. We characterize noise robustness of the gradient scheme in terms of spectral properties of signless Laplacians of the interaction graph. 
 We then demonstrate that a plug-in estimator based on the estimated skills achieves state-of-art performance on a number of real-world datasets. Our results have implications for rank-one matrix completion problem in that gradient descent can provably recover $W \times W$ rank-one matrices based on $W+1$ off-diagonal observations of a connected graph with a single odd-cycle.
%
\end{abstract}

\section{Introduction}
Crowdsourcing is being utilized as a scalable approach for rapidly collecting annotated data for a diverse set of applications, including image recognition and natural language processing. Due to the high variability of worker skills, crowd-sourcing solutions aggregate inputs from a large number of workers for each task. In this context many aggregation methods 
that incorporate worker quality have been proposed. Recent works \cite{BeKo14:NIPS,Sze15:MSc} have investigated the importance of having a precise knowledge of skill quality for accurate prediction of ground-truth labels.  
On the other hand, in practice, worker-task assignments are sparse and irregular due to the arbitrary,
and uncontrolled availability of workers on crowdsourcing platforms leading to difficulties in estimating worker skill-level.

Motivated by these findings we develop worker skill estimation methods for the symmetric single-coin model \cite{DaSke79} and focus on binary classification tasks \footnote{
In our experiments, we will also consider the multi-class case using a one-vs-all encoding of multiclass labels and by assuming a common skill level for any worker across the labels.
While this reduction is arguably limited,
the limitations of this reduction will be seen not to degrade performance (as compared
to the performance of competing methods) agnostic to the nature of worker-task assignments.
It remains for future work to consider further alternatives to our reduction.
}. Our data is a sparse $W \times T$ worker-task interaction matrix with $W$ workers and $T$ binary tasks, the
$ij$th component filled by the label provided by the $i$th worker for the $j$th task when the worker indeed provided such a label. As in \cite{DaSke79}, we assume that the workers independently label tasks. Further, we assume that each worker's skill is parameterized by a single skill parameter, which determines the probability of the worker flipping the sign of the true label. 
Unlike \cite{DaSke79}, we make no further assumptions either on task priors or independence of ground-truth labels across tasks. 

Our goal is to characterize structural properties of the sparsity pattern of the worker-task interaction matrix required for accurately estimating worker skills. On the one hand worker skills can be consistently estimated, in general, if each task is labeled by all workers (no missing data), while skills cannot be estimated if each task is only labeled by a single worker (sparse setting). Our goal is motivated by the need to understand the sparsest setup that would lead to consistent skill estimates. One of the key contributions of this paper is in identifying such a condition. Surprisingly, it turns out that a necessary and sufficient condition can be stated in terms of the properties of the so-called \emph{(worker) interaction graph}. 
Nodes in this graph are associated with workers. An edge indicates whether or not the associated nodes (workers) participated in a common task. 
Further, the ``limiting interaction graph'' is one when two workers are connected if they participated in infinitely many common tasks.
We show that skills are asymptotically identifiable if and only if the limiting interaction graph is irreducible and has odd-cycles. 

We next propose 
a statistically consistent method for skill estimation. We formulate a weighted rank-one minimization function, 
where the objective is to minimize the difference between expected correlation (product of the skills) and the observed correlations. The objective function is non-convex and we develop a 
 gradient descent scheme by recursively updating skill levels to estimate worker skills. 
An important aspect of the lack of supervision is that covariances corresponding to each worker cannot be measured. Consequently, the problem cannot in general be reduced to conventional rank-one approximation and must be dealt with in its full generality. 

We show in the noiseless case for aperiodic, irreducible graphs, the equilibrium point of the gradient descent algorithm is unique and matches the skill vector. We then develop error bounds for noisy correlation data. 
We characterize bounds for skill estimation error in terms of spectral properties of the interaction graph.

We then test our proposed approach on both synthetic and real-world data. For estimating prediction accuracy we use a plug-in estimator based on estimated skills. We also provide a simple extension of the method to the multi-class case. We demonstrate on several real-world experiments that our proposed approach achieves state-of-art performance on both binary and multi-class datasets. 

\newcommand{\normsm}[1]{\|#1\|}
\newcommand{\norm}[1]{\left\|#1\right\|}
\textbf{Notation and conventions}: 
The set of reals is denoted by $\R$, the set of natural numbers which does not include zero is denoted by $\N$. 
For $k\in \N$, $[k] \defeq \{1, \dots, k \}$.
Empty sums are defined as zero. We will use $\PP$ to denote the probability measure over the measure space
holding our random variables, while $\mathbb{E}$ will be used to denote the corresponding expectation operator.
For $p\ge 1$, we use $\norm{v}_p$ to denote the $p$-norm of vectors. Further, $\norm{\cdot}$ stands for the $2$-norm. 
The cardinality of a set $S$ is denoted by $|S|$. Proofs of new results, missing from the main text are given in the appendix.
\vspace{-0.1in}
\section{Problem Setup}
\vspace{-0.05in}
We consider binary crowd sourcing tasks where a set of workers provide binary labels for a large number of items. Let $W\in \N$ be a fixed positive integer denoting the number of workers.  A problem instance $\theta \doteq (s,A,g)$ 
is given by a skill vector 
$s = (s_1,\dots,s_W)\in [-1,1]^W$, 
the worker-task assignment set $A\subset [W]\times \N$
and the vector of ``ground truth labels'' $g\in\pms^{\N}$. 

When $A\subset [W]\times [T]$ for some $T\in \N$, we say that $\theta$ is a finite instance with $T$ tasks, 
otherwise $\theta$ is an infinite instance. 
We allow infinite tasks to be able to discuss asymptotic identifiability.
The set of all instances is denoted by $\Theta$, 
the set of finite instances
with $T$ tasks is denoted by $\Theta_T$.
The (worker) interaction graph is a graph $G= ([W],E)$ over $[W]$ with $i,j\in [W]$ connected ($(i,j)\in E$) in $G$ if there exists some task $t\in \N$ such that both $(i,t)$ and $(j,t)$ are an element of $A$.

The problem in label recovery with crowd sourcing  
is to recover the ground truth labels $(g_t)_t$ 
given observations $(Y_{w,t})_{(w,t)\in A}$, a collection of $\pm 1$-valued 
random variables such that $Y_{w,t} = Z_{w,t} g_t$ for $(w,t)\in A$, 
where $(Z_{w,t})_{(w,t)\in A}$ is a collection of 
independent random variables that satisfies $\E[Z_{w,t}]=s_{w}$.


A (deterministic) inference method takes the observations 
$(Y_{w,t})_{(w,t)\in A}$ and 
returns a real-valued score for each task in $A$;
the signs of the scores give the label-estimates.
Formally, we define an inference method as a map 
$\gamma: \pms^A \to \mathbb{R}^\N$, 
where given $Y\in \pms^A$, $\gamma_t(Y)$, 
the $t$th component of $\gamma(Y)$ is the score inferred for task $t$ given the data $Y$. 
When important, we will use the subindex $\theta$ in $\PP_\theta$ to denote the dependence of the probability 
distribution over the probability space holding our random variables. We will use $\E_\theta$ to denote
the corresponding expectation operator.

The average loss suffered by an inference method $\gamma$ on the first $T$ tasks of an instance $\theta$ 
is
\[
L_T(\gamma;\theta) = 
	\frac1T\, \E_{\theta}\Bigl[ 
     \textstyle	\sum_{t=1}^T\one{\gamma_t(Y) g_t \leq 0} \Bigr]\,.
\]
\subsection{Two-Step Plug-in Approach}
We propose a two step approach based on first estimating the skills and then utilizing a plug-in classifier to predict the ground-truth labels. The motivation for a two-step approach stems from existing results that characterize accuracy in terms of skill estimation errors. We recall some of these facts here for exposition. 

For future reference, define the log-odds weighted majority vote parameterized by parameter vector $\alpha \in (-1,1)^W$: 
\[
\gamma_{t,\alpha}(Y) = \sum_{(i,t)\in A}v(\alpha_i) Y_{i,t},\,\, \mbox{where}\,\,\,v(\alpha) = \log{\frac{1+\alpha}{1-\alpha}}.
\]
\cite{NiPa81:WM} showed that the optimal decision rule,
which minimizes 
the probability of making error $\Prob{\gamma_{t,s}(Y) g_t \leq 0}$
individually for every $t\in \N$,
 is a weighted majority vote with parameter $\alpha = s$, giving the weights $v_i^* = v(s_i)$. 
%
%
%
We denote this optimal decision rule as $\gamma^*$. 
%

When skills are known, \cite{BeKo14:NIPS} provide an upper error bound, as well as a asymptotically matching lower error bounds in terms of the so called {\it committee potential}.
When skills are only approximately known, \cite{Sze15:MSc,BeKo14:NIPS} also show that similar results can be obtained:
\begin{lemma}
\label{Lem:predictionerror}
For any $\epsilon > 0$, the loss with estimated weights $\hat{v}_i = v(\hat s_i)$ satisfies 
\[
\frac1T\, \E_{\theta}\Bigl[ 
    	\sum_{t=1}^T\one{\gamma_{t,\hat s}(Y) g_t \leq 0} \Bigr] \leq \frac1T\, \E_{\theta}\Bigl[ 
    	\sum_{t=1}^T\one{\gamma^*(Y) g_t \leq \epsilon} \Bigr] + \PP_{\theta}(\|v^*-\hat{v}\|_1 \geq \epsilon)\,.
\]
\end{lemma}
We can express the error term (2nd term on the RHS) in the above equation in terms of the
multiplicative norm-differences in the skill estimates (see \cite{BeKo14:NIPS}).
\begin{lemma} \label{lem:errorV}
Suppose $\frac{1+ \hat s_i}{1+s_i}, \frac{1- \hat s_i}{1-s_i} \in [1-\delta_i, 1+\delta_i]$ then 
$|v(s_i) - v(\hat s_i)| \leq 2|\delta_i|$.%
\end{lemma}
These results together imply that a plug-in estimator with a guaranteed accuracy on the skill levels in turn leads to a bound on the error probability of predicting ground-truth labels. Therefore, we focus on the skill estimation problem in the sequel. 
\section{Weighted Least Squares Estimation} \label{sec:WLS}
In this section, we propose an asymptotically consistent skill estimator with missing data. By missing data, 
we not only mean that  only a subset of workers provide labels for a given task, but more importantly we mean that 
the interaction graph is not a click.

Recall that given an instance $\theta=(s,A,g)$, 
the data of the learner is given in the sparse matrix
$(Y_{i,t})_{(i,t)\in A}$ which is a collection of independent binary random variables such that $Y_{i,t}=g_tZ_{i,t}$ and $s_i=\mathbb{E}(Z_{i,t})$.
Define $N\in \N^{W\times W}$ to be the matrix whose $(i,j)$th entry gives how many times workers $i$ and $j$ labeled the same task:
\begin{align*}
N_{ij}=|\{t\in \N:(i,t),(j,t)\in A\}|\,.
\end{align*}
Note that the there is an edge between workers $i$ and $j$ in the interaction graph, denoted by $G=([W],E)$, 
exactly when $N_{ij}>0$. That is, $(i,j)\in E$ if and only if $N_{ij}>0$. When $A$ is infinite, $N_{ij}$ may be infinite.

Let $\theta$ be a finite instance.
When $(i,t),(j,t)\in A$,
since $g_t^2=1$, by our independence assumptions, $\EE{ Y_{i,t},Y_{j,t} } = s_i s_j$.
This motivates estimating the skills using 
\begin{equation}
\label{Eq:Objective}
\tilde{s}=\argmin_{x\in[-1,+1]^W}\frac{1}{2}\sum_{(i,t),(j,t)\in A}(Y_{i,t}Y_{j,t}-x_{i}x_{j})^2\,.
\end{equation}
Assume now that $\theta$ is a finite instance.
Define
$C_{ij} \defeq s_i s_j$ and let
\begin{equation*}
\tilde{C}_{ij}=\frac{1}{N_{ij}}\textstyle \sum_{(i,t),(j,t)\in A}\, Y_{i,t}Y_{j,t}\,.
\end{equation*}
An alternative interpretation of the objective in Eq.~\eqref{Eq:Objective} is given by the following result:
\begin{lemma}\label{lem:equiv}
Let $L:[-1,1]^W \to [0,\infty)$ be defined by 
$L(x) = \sum_{(i,j)\in E}N_{ij}(\tilde{C}_{ij}-x_ix_j)^2$. 
The optimization problem of Eq.~\eqref{Eq:Objective} is equivalent to the optimization problem
$\argmin_{x\in[-1,+1]^W}L(x)$.
\end{lemma}
The above result shows that our estimation problem can alternatively be described as a (sparse) weighted rank-one approximation problem. Such problem are in general hard \cite{GiGli11:NPhardness}. However, our data has special structure, which may allow one to avoid the existing hardness results.

Our theoretical analysis, which we defer to Sec.~\ref{sec:theory}, will establish that the skills are asymptotically identifiable 
in an infinite instance $\theta$ if and only if the so-called limiting interaction graph, in which two workers are connected if and only if $N_{ij}=\infty$, 
is irreducible and has an odd-cycle.

\noindent {\bf Remark:} An important aspect of the lack of supervision is that the diagonal elements $N_{ii}$ are zero since otherwise this would imply direct measurements of worker-skills. Consequently, the matrix $N$ is in general full-rank and the problem cannot be reduced to a standard rank-one approximation problem. 

\subsection{Plug-in Projected Gradient Descent}
To solve the weighted least-squares objective, we propose a \emph{Projected Gradient Descent} (PGD) algorithm (cf. \cref{algo:PGD}). At each step we sequentially update the skill level based on following the negative gradient:
\begin{align*}
\tilde{s}^{t+1}_i=&s^{t}_i+\gamma\sum_{(i,j)\in E} N_{ij} (\hat{C}_{ij}-s^{t}_is^{t}_j)s^t_{j}\\
s^{t+1}_i=&P(\tilde{s}^{t+1}_i),
\end{align*}
where $P(\cdot):\mathbb{R}\rightarrow\left[-1+\frac{\tau}{\sqrt{\mathcal{N}_i}},+1-\frac{\tau}{\sqrt{\mathcal{N}_i}}\right]$ is a projection function, $\gamma>0$ is the step size; 
%
$N_i=|\{t:(i,t)\in A\}|$ is the number of tasks labeled by worker $i$ and $\tau>0$ is a tuning parameter.
We further use 
a weighting function $B: \mathbb{R}^+ \rightarrow \mathbb{R}^+$, which can be simply the identity,
but other weighting functions are also permitted as long as $B(0)=0$ and $B(\alpha) \geq 0$.  

\begin{wrapfigure}{r}{0.5\textwidth}
	\begin{minipage}{.49\textwidth}
		\begin{algorithm}[H]\label{algo:PGD}
			\caption{Plug-in Projected Gradient Scheme}
			\begin{algorithmic} 
				\STATE \textbf{Input}: $N$, $Y=\{Y_{i,t}\}_{(i,t)\in A}$, $\eta,\tau>0$.
				\STATE  $x_i\sim U[-1,1]$
				\STATE  $\tilde{C}_{ij} \leftarrow \frac{1}{N_{ij}} \sum_{(i,t),(j,t)\in A} Y_{i,t} Y_{j,t}$, $\forall (i,j)$ s.t. $N_{ij}>0$.
				\REPEAT
				\FOR{$i=1,\ldots,W$}
				\STATE $x_i \leftarrow x_i+2\eta\sum_{j=1,\ldots,W}N_{ij}\tilde{C}_{ij} x_{j}$
				\STATE $~~~~~~~~~~~~~-2\eta\sum_{j=1,\ldots,W}N_{ij}x_i x_j^2$
				\STATE $x_i \leftarrow \max\{x_i,1-\frac\tau{\sqrt{N_i}}\}$
				\STATE $x_i \leftarrow \min\{x_i,-1+\frac\tau{\sqrt{N_i}}\}$
				\ENDFOR
				\UNTIL{$x$ converges}
				\STATE $\hat{s}\leftarrow \sgn(\sum_{i=1}^W x_i) x$
				\FOR{$t=1,\ldots,T$}
				\STATE $\hat{Y}_t \leftarrow \sum_{i=1}^W  Y_{i,t}\,\log \frac{1+\hat{s}_i}{1-\hat{s}_i}$
				\ENDFOR
				\STATE \textbf{return} $(\hat{Y}_t)_{t\in [T]}$
			\end{algorithmic}
		\end{algorithm}
		\vspace{-.70cm}
	\end{minipage}%
\end{wrapfigure}

The purpose of the projection is to stay away from the boundary of the hypercube, where the log-odds function is changing very rapidly.
Our justification is that skills close to one have overwhelming impact on the plug-in rule 
and since the skill estimates are expected to have an uncertainty proportional to $\tau/\sqrt{N_i}$ with probability $\mathrm{const}\times e^{-\tau}$,
 there is little loss in accuracy in confining the parameter estimates to the appropriately reduced hypercube.
 While in principle one could tune this parameter, we use $\tau=1$ in this paper. 
Finally, the PGD algorithm (or any other algorithm) without further information can only identify the skill vector to be one of the antipodal possibilities $(\pm s)$. We assume that $\sum s_i > 0$ and assign signs accordingly.

\noindent {\bf Remark:} Note that we could employ a number of different weighting functions. Our theoretical analysis shows that any weighting function satisfying positivity and $B(0)=0$ leads to convergence in the noiseless setting. 
\section{Theoretical Results} \label{sec:theory}
In this section we derive theoretical results to shed light on the fundamental structural properties required of the assignment matrix to ensure asymptotic identifiability with missing data. Subsequently, we analyze convergence properties of the PGD algorithm for noiseless and noisy cases. 

\subsection{Identifiability}
We frame identifiability in terms of whether or not the average regret converges to zero. 
The average regret of an inference method $\gamma$ for instance class $\Theta_{s,A}$ that 
contains all instances that share the same skill vector $s$ and assignment matrix $A$ as
\[
\oR_T(\gamma;\Theta_{s,A}) = \sup_{\theta\in \Theta_{s,A}} L_T(\gamma;\theta) - L_T(\gamma^*(s,A);\theta)\,.
\]
While this regret is worst-case for the ground-truth, it is instance specific as far as the skills $s$ and the task-worker assignment $A$ are concerned. Also note that the regret formulation bypasses the objective of estimating the ground-truth. Rather the goal is to match cummulative loss against a competitor\footnote{This perspective also arises in related sequential sensor selection problems\cite{hanawal17a} where one must learn to make decisions in the absence of ground-truth.}. 

To deal with the asymptotic tasks case we characterize connectivity based on whether two workers are together active  finitely many times only, or infinitely many times. As alluded to before, the limiting interaction graph has an edge between two workers if and only if $N_{ij}=\infty$. Alternatively, when $N_{ij}=\infty$ we will say that workers $(i,j)$ are \emph{connected} in $A$, otherwise we say that they are \emph{disconnected} in $A$.

Let $\Theta'\subset \Theta$ be a ``truth-complete'' set of instances: 
That is, for any $\theta = (s,A,g)\in \Theta'$, $\Theta_{s,A}\subset \Theta'$.
Truth-completeness expresses that there is no a priori information about the unknown labels.
An inference method is said to be \emph{consistent} for a truth-complete instance set $\Theta'\subset \Theta$ 
if for any $(s,A)$ such that $\Theta_{s,A}\subset \Theta'$, $\limsup_{T\to\infty} \oR_T(\gamma;\Theta_{s,A})=0$.

Given $\Theta'\subset \Theta$, 
we let $S(\Theta') =  \{ s\in [-1,1]^W\,:\, (s,A,g)\in \Theta' \}$ be 
the set of skill vectors underlying $\Theta'$.
For a skill vector $s\in [-1,1]^W$ we let $P(s) = \{ i\in [W]\,:\, s_i>0 \}$ to be the set of workers
whose skills are positive and we let 
$\mathcal{P}(s) = \{ P(s), P(-s) \}$ be the grouping of workers into workers with positive and negative skills.
Note that workers with zero skill are left out.
Finally, we define $\Theta_A =  \{ (s,A,g)\,:\, s\in [-1,1]^W, g\in \pms^\N \}$ as the set of all instances where the assignment set is given by $A$.

We have the following theorem: 
\begin{theorem}[Characterization of learnability] \label{thm:learnability}
Consider an assignment set $A$ such that the limiting interaction graph $G = G(A)$ of $A$ has a single component.
Let $\Theta' \subset \Theta_A$ be a truth-complete set of instances over $A$.
Then $\Theta'$ is learnable if and only if the following hold:
\begin{enumerate}[(i)]
\item For any $s,s'\in S(\Theta')$ such that $|s| = |s'|$ and $\mathcal{P}(s) = \mathcal{P}(s')$, it follows that $s=s'$;
\label{thm:c1}
\item There exists an odd cycle%
\footnote{An odd cycle is a cycle with an odd number of vertices.} in $G$.
\label{thm:c2}
\end{enumerate}
\end{theorem}
The forward direction of the theorem statement hinges upon the following result:
\begin{lemma}\label{lem:asylearn}
For any $g\in \pms$, $s\in [-1,1]^W$ and an assignment set with a single-component limiting interaction graph $G$ which has
at least one odd cycle,
there exists a method to estimate $|s|$ and $\mathcal{P}(s)$.
\end{lemma}
The reverse implication in the theorem statement follows from the following result:
\begin{lemma}\label{lem:asylearn1}
Assume that the lengths of all cycles in $G$ are even. Then there exists $s,s'\in [-1,1]^W$, $s\ne s'$ such that $C_{ij}=s_is_j = s_i's_j'$.
\end{lemma}
Theorem~\ref{thm:learnability} (see proof), suggests one way to ensure learnability is to assume that $\sum_{i\in[W]}s_{i}>0$, which we do so in this paper. 
\subsection{Convergence of the PGD Algorithm}
The previous section established that for identifiability 
we require the limiting interaction graph to be connected and must have an odd-cycle. We will now show that PGD under these assumptions converges to a unique minimum for both the noisy and noiseless cases;
by the latter we mean that in the loss $L$ of \cref{lem:equiv}, we set $\tilde C_{ij} = C_{ij}=s_is_j$ for $(i,j) \in E$.
Note that the odd-cycle condition together with that $G$ is connected gives that the worker-interaction count matrix $N$ is irreducible and aperiodic.
We show that in this case the loss has a unique minima and the PGD algorithm recovers the skill-vector. 
\begin{theorem} \label{thm:noiselessPGD}
Suppose the worker-interaction matrix $N$ is irreducible, aperiodic and the components of the skill vector are non-zero. Then the PGD Algorithm of Sec~\ref{algo:PGD} for the noiseless case converges to the global minima. It follows that skill estimates are asymptotically consistent if the limiting interaction graph with weights $\rho_{ij}=\lim_{T\rightarrow \infty} N_{ij}(T)/T >0$ is irreducible and aperiodic.
\end{theorem}
The proof of the result is based on analyzing the critical points of the loss $L$ underlying the PGD algorithm. 
Specifically, we wish to verify whether or not there exists a vector $x \neq s$ such that, for each $i=1, \ldots,W$, we have
\begin{equation} \label{maineq} 
\sum_{j=1}^W N_{ij} (x_i x_j - s_i s_j) x_j = 0. 
\end{equation}
We argue that when worker-interaction matrix $N=[N_{ij}]$ is irreducible and does not contain even cycles, the only two points that satisfy this equation are $x=s$ and $x=-s$. We then rule out the incorrect equilibrium point by invoking our prior assumption that $\sum_i s_i > 0$. Finally, by means of the second order conditions we check that the critical points are indeed minima. To see this we check the Hessian $P(x) \defeq \nabla^2 L(x)$ of $L$. First,  trivial algebra gives:
\begin{equation} P_{ii}(x) =  \sum_{j=1}^n 2 N_{ij} x_i^2,\,\, P_{ij}(x)  =  4 N_{ij} x_i x_j - 2 N_{ij}s_is_j
\end{equation}
Notice that $P(s)=[P_{ij}(s)] = 2\text{diag}[s_i]P(1)\text{diag}[s_i]$. Positive definiteness of $P(s)$ follows from the fact that $P(1)$ is unsigned Laplacian matrix since the diagonal is the sum of the off-diagonal entries, which are all positive. The fact that unsigned Laplacians are positive definite for non-bipartite graph follows from results of \cite{desai94}.

Note that for the noiseless case, the theorem imposes few restrictions on the interactions in terms of number of tasks per worker, the total number of tasks, or whether task assignments can be asynchronous. Indeed, interactions could involve only two workers for each task and yet PGD converges to the skill-vector.

We will now extend these results to the noisy case. 
There are two alternative proofs for our result. Although we can invoke the implicit function theorem we leverage local strong convexity of the gradient to determine the size of the perturbation suffered due to noise. To this end, we consider the equilibrium points of the PGD for the noisy case again: 
\begin{equation} \label{noisemaineq} 
\sum_{j=1}^W N_{ij} (x_i x_j - \tilde C_{ij}) x_j = \sum_{j=1}^W N_{ij} (x_i x_j - s_is_j + \delta_{ij}) x_j = 0 \implies \sum_{j=1}^W N_{ij} (x_i x_j - s_is_j) x_j= \sum_{j=1}^W N_{ij}\delta_{ij}
\end{equation}
where $\Delta=[\delta_{ij}]$ is a perturbation encountered due to noisy estimate of the correlation between nodes $i$ and $j$. We now state a theorem in terms of the graph-theoretic properties of the worker-interaction matrix. 
We then have the following result:
\begin{theorem} \label{thm:noisyPGD}
Suppose the worker-interaction matrix satisfies the assumptions in Theorem 2, Then for each $\epsilon \in (0,1)$ there exists a constant $c_{\epsilon} > 0$ with the following property: if $\Delta \in \R^{W \times W}, x \in \R^W$ satisfies Eq.~\ref{noisemaineq},  $|x_i| \geq \epsilon$ and the noise matrix satisfies $\|\Delta\|_2 \leq c_{\epsilon}$, where $c_{\epsilon}$ is some constant depending only on $\epsilon$
then 
\begin{equation*} 
\|x - s\|_2 \leq \frac{\|N\|_F\|\Delta\|_2}{s_{\min}^2 \sigma_{\min}(P(1))},\,\,\, \mbox{where}\,\,\, s_{\min}=\min_i |s_i|.
\end{equation*}
\end{theorem}
Numerator in the above expression depends on the number of tasks but does not pose a problem since the denominator also scales with the number of counts and essentially cancels out the scaling. 
%


\section{Experimental Results}
{\bf \textsc{Synthetic Experiments:}}
We will experiment with the impact of noise, graph-size, skill distribution and different weighted graphs on synthetic data. Experiments with graph size and skill distribution appear in supplementary material. Here we highlight robustness of PGD to noise and the superiority of using the identity weights (see Lemma~\ref{lem:equiv}) over other weights. We consider two types of graphs here with 11 nodes (workers). Graph $G_1$ is a clique worker-worker interaction graph and Graph $G_2$ is a star-graph with an odd-cycle of length 3.

\begin{figure}[h]
\vspace{-0.1in}
\centering
\subfigure{\label{subFig:clique}}\addtocounter{subfigure}{-1}
\subfigure[Clique.]{
\includegraphics[width=0.45\textwidth]{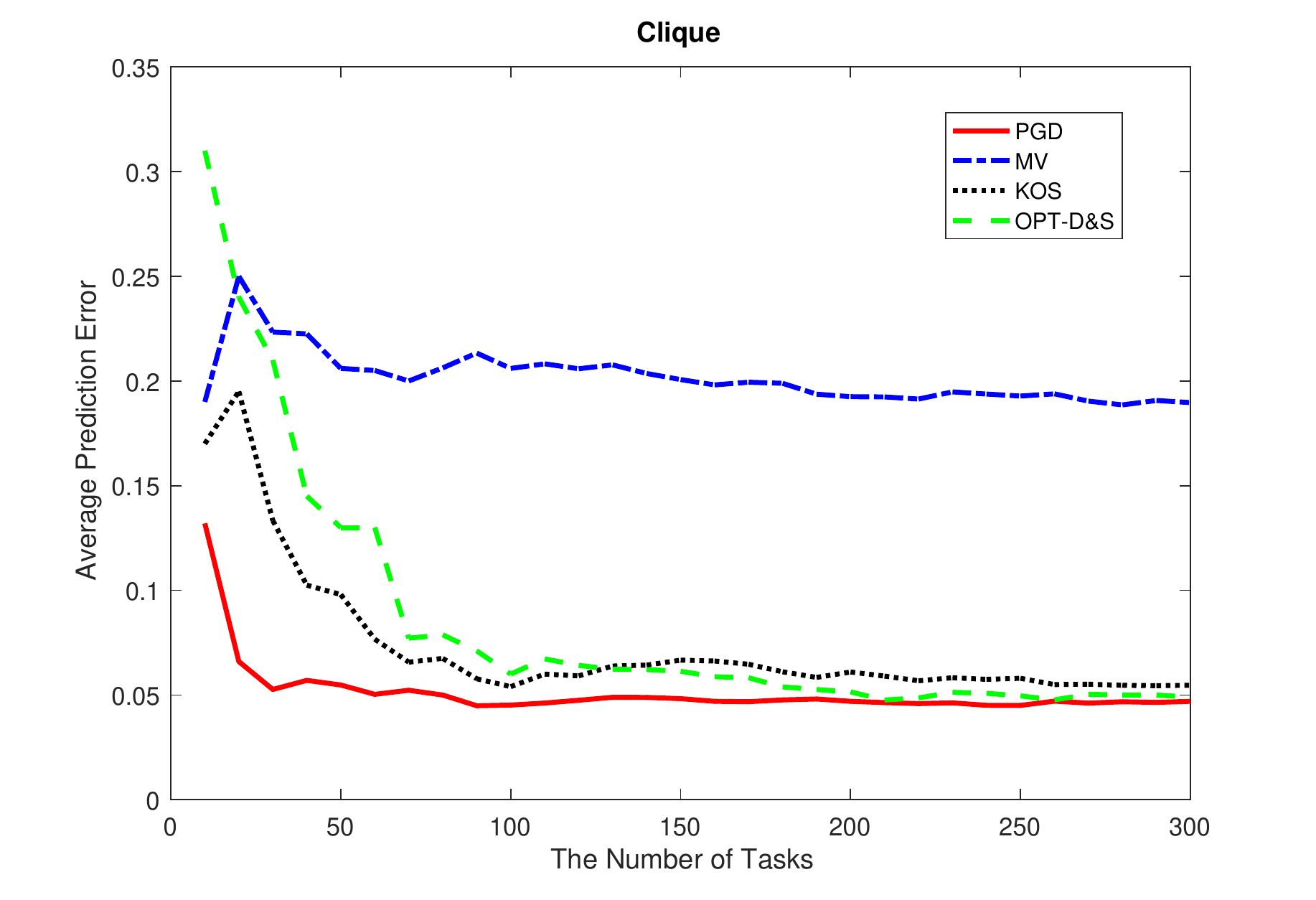}}
\subfigure{\label{subFig:star}}\addtocounter{subfigure}{-1}
\subfigure[Star graph with $3$-cycle.]{
\includegraphics[width=0.45\textwidth]{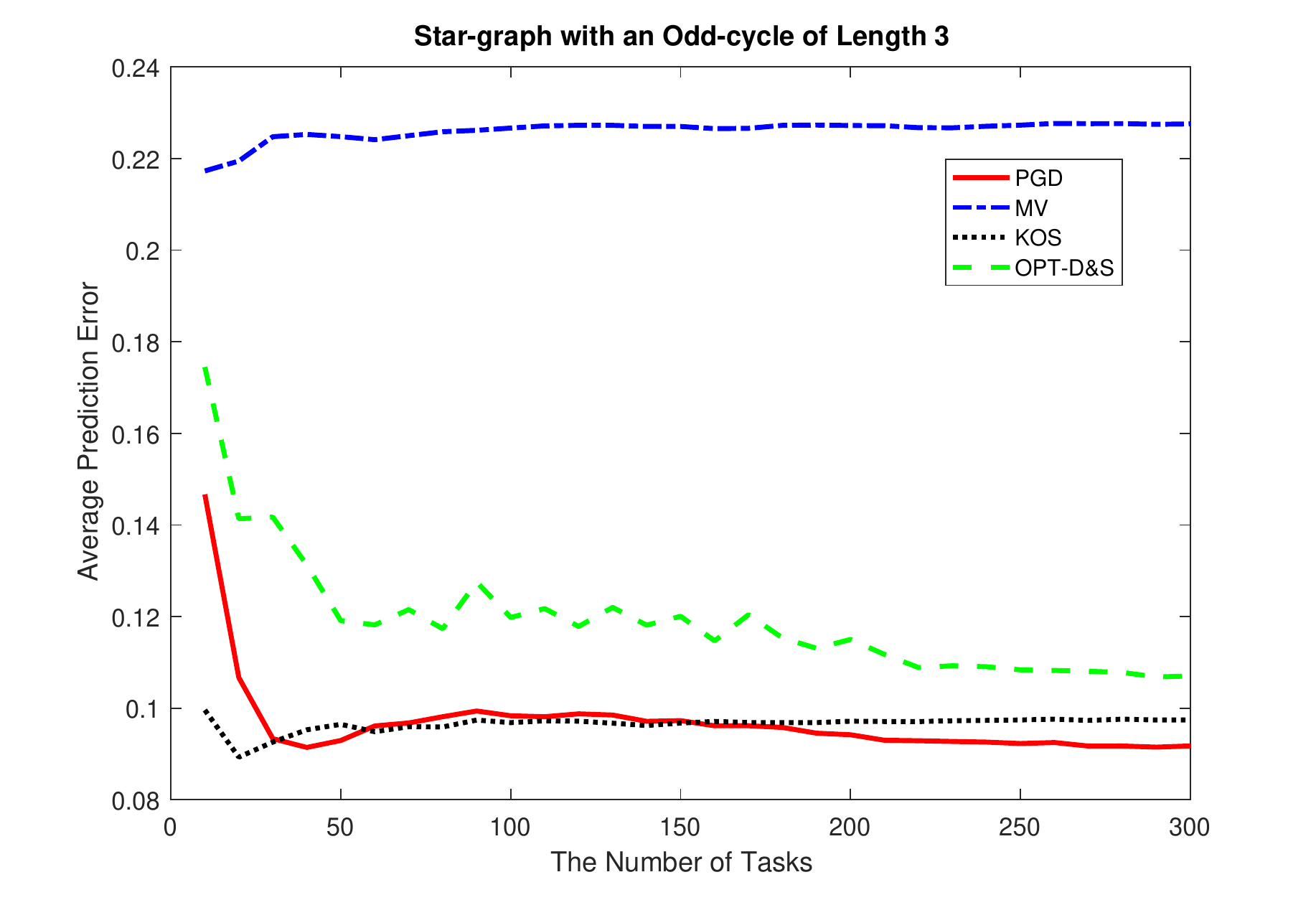}}
\label{Fig:comparison}
\vspace{-0.05in}
\caption{\small Illustrative comparisons of prediction performance two graph types. Only mean values are plotted for exposition. For the clique, the standard deviation values with 10 tasks were $0.1647$, $0.1524$, $0.2163$, and $0.2514$ for PGD, MV, KOS, and OPT-D\&S respectively; and with 300 tasks they were $0.0120$, $0.0172$, $0.0172$, and $0.0168$ for PGD, MV, KOS, and OPT-D\&S respectively. For the star-graph the standard deviations for 10 tasks were $0.2144$, $0.0815$, $0.1058$, and $0.2565$ for PGD, MV, KOS, and OPT-D\&S respectively and for 300 tasks they were $0.0181$, $0.0210$, $0.0184$, and$0.0304$. Standard deviations decrease with growing number of tasks.}
\vspace{-0.1in}
\end{figure}

\noindent
{\it Noise Robustness:} To see the impact of noise We vary the noise level by increasing the number of tasks, which in turn reduces the error in the correlation matrix. 
Tasks are randomly assigned to binary classes $\pm{1}$ with total number of tasks ranging from $30$ to $300$. Skills are randomly assigned on a uniform grid between $0.8$ and $-0.3$. 
We compare the average prediction error $PE=\frac{1}{T}\sum_{t=1,\ldots,T}1\{\hat{Y}_t\neq g_t\}$ with the majority voting algorithm, the KOS algorithm \cite{Karger2013}, and Opt-D\&S algorithm \cite{Zhang2014}.
Each algorithm is averaged over 300 trials on each dataset. The average prediction errors are presented in Figure~\ref{Fig:comparison}. As the number of tasks grows, the average prediction error of PGD algorithm decreases. In terms of comparison, PGD algorithm appears to be robust to graph-structure (and hence missing data), while the closest competitor (OPT-DS) has significant performance degradation on sparse graphs. 
%
%

\noindent{\it Graph Weights:} We argued in Sec.~\ref{sec:theory} that PGD algorithm converges to the global optimal for any non-negative weights. It is interesting to consider the behavior with different choices. \cite{dalvi_aggregating_2013} has suggested using $B(N_{ij})=N_{ij}^2$, while we use $N_{ij}$. Another possibility is to use binary weights. 
\begin{wrapfigure}{r}{0.5\textwidth}
\vspace{-0.1in}
\makeatletter\def\@captype{table}\makeatother
  \caption{\small Prediction errors for different Weightings}
\label{Tab:weighting}
\centering
\scalebox{0.65}{
\begin{tabular}{c|c|c|c}
\hline
Worker type  \\ Assigned most tasks &  $[N_{ij}>0]$  & $B(N_{ij})=N_{ij}$ & $B(N_{ij})=N_{ij}^2$\\ \hline
Spammers & $0.33\pm{0.03}$  & $0.33\pm{0.03}$ & $0.55\pm{0.17}$\\ \hline
Positive skill workers& $0.17\pm{0.06}$ & $0.09\pm{0.02}$ & $0.09\pm{0.02}$\\ \hline
\end{tabular}}
\end{wrapfigure}
We iteratively run PGD $10$ times for each weighing function with $T=300$ tasks for different types of task assignements. If $N_{ij}$'s are all equal, these choices produce identical results. We consider two cases: (a) Spammers are assigned a majority of tasks; (b) Positively skilled workers are assigned most tasks. The prediction errors are compared in Table~\ref{Tab:weighting}. Note that quadratic weighting is quite bad in this case because it tends to ignore positively skilled workers. On the other hand unweighted case does not accurately estimate spammers and also results in poor choice.

%
%


\begin{table}[h]
\caption{Benchmark Datasets with Prediction Errors for Different Methods.}
\label{Tab:realdata}
\scalebox{0.75}{
\begin{tabular}{c|c|c|c|c|c}
\hline
Datasets & Tasks & Workers & Instances & Classes & Sparsity level\\ \hline
RTE1     & 800  &  164    & 8000    &   2 &0.0610\\ 
Temp     & 462   & 76      &4620      &   2 &0.1316\\ 
Dogs      & 807   & 109     & 8070    &   4&0.0917\\ 
Web      & 2665    & 177      & 15567    &   5&0.0033\\ 
\hline
\end{tabular}}
\quad
\scalebox{0.85}{
\begin{tabular}{c|c|c|c|c}
\hline
Data & MV & Opt-D\&S &KOS & PGD\\ \hline
RTE1   & 0.1031        & 0.0712       &  39.75 &\textbf{0.07}\\ 
Temp   & 0.0639       &  0.0584        & 0.0628  &\textbf{0.054}\\ 
Dogs    & 0.1958  &  0.1689          & 31.72  &\textbf{0.1660}\\ 
Web    & 0.2693  &  \textbf{0.1586}             & 42.93  &0.1623\\ 
\hline
\end{tabular}}
\vspace{-0.16in}
\end{table}
{\bf \textsc{Benchmark Dataset Experiments:}}
We illustrate the performance of PGD algorithm against state-of-art algorithms described before. Each algorithm is executed on four data-sets, i.e. RTE1 \cite{Snow08}, Temp \cite{Snow08}, Dogs \cite{Deng2009}, and WebSearch \cite{Zhou2012}. A summary of these data-sets is presented in Table~\ref{Tab:realdata}.
 RTE1 and Temp data-sets have binary labels where our algorithm could be directly applied to. For the multiclass data-sets (i.e., Dogs and Web), We iteratively run our algorithm with one-vs-rest strategy. A score function is calculated at the end of each iteration as follows
$score(k)=\sum_{(i,t)\in A}\log{\frac{1+s_i}{1-s_i}}\bm{1}(Y_{i,t}=k)$,
where $k\in \mathcal{K}$ is the class index and $\bm{1}(\cdot)$ is a $\pm 1$ indicator. Then we predict the label by finding the maximum of the score function. The results  summarized in Table~\ref{Tab:realdata}, shows that proposed PGD algorithm outperforms the state of the art algorithms and is comparable to Opt-D\&S on Web. We attribute the small excess error to the fact that Opt-D\&S estimates the multi-class confusion matrix while we do a one-vs.-all.


\vspace{-0.05in}
\section{Related Work}
To place our results in the literature, 
recall that our approach is to estimate the unknown labels using weighted majority 
using a log-odds based plug-in estimator that uses the estimated skill of the workers,
while skill estimation uses a weighted rank-one approximation of the  
empirical worker-worker agreement matrix. 
Like many other works we adopt the generative model of \cite{DaSke79} for deriving our approach.
In the context of this model,
\cite{EHR11} proposed the use of weighted majority voting,
but log-odds based weights were first analyzed by
\cite{zhou2013error}.
\cite{BeKo14:NIPS} and \cite{Sze15:MSc} further studied the properties of the corresponding plug-in estimates
and we will build on their results.

Weighted rank-one approximation has been used by \cite{dalvi_aggregating_2013} and in many ways
this work is the closest to ours.
Like we, they are also motivated by the desire to derive methods that work with sparse,
and non-regular worker-task assignments, which are often observed in practice (due to the arbitrary,
and uncontrolled availability of workers on crowdsourcing platforms).
However, their loss function and approach differs from ours:
In particular, they consider two loss functions.
Using our notation, the first loss they consider is 
$L_{\mathrm{DDKR}}^{(1)}(x) = \sum_{ij} N_{ij}^2( C_{ij}-x_i x_j)^2$, while the second loss
is 
$L_{\mathrm{DDKR}}^{(2)}(x) = \sum_{ij:N_{ij}>0} (C_{ij} - x_i x_j)^2$. 
The second loss is simply the unweighted loss, while the first 
loss differs from ours because the individual error terms are weighted by $N_{ij}^2$. They derive the first loss 
by considering the unnormalized worker-worker agreement matrix whose $(i,j)$th entry is $A_{ij} \defeq \sum_{t} Y_{i,t} Y_{j,t}$:
The expectation of this is $N_{ij} s_i s_j$, which led to to consider the loss 
$L_{\mathrm{DDKR}}^{(1)}(x) = \sum_{ij} (A_{ij} - N_{ij} x_i x_j)^2$, which, using $A_{ij} = N_{ij} C_{ij}$ leads to the first expression shown for $L_{\mathrm{DDKR}}^{(1)}$.
They provide an error bound for an algorithm that uses the top eigenvectors of $A$ and $N = (N_{ij})$ in terms of the spectral properties of $N$: A large spectral gap between the first two top eigenvalues of $N$ 
is shown to give a more accurate estimates of the skills. In their experiments, they use the spectral algorithm to initialize
and alternating projection algorithm.

As noted, our loss uses a statistically justifiably choice of weighting:
In particular, the variance of $C_{ij}-s_i s_j$, which is the expectation of the square of this difference
 is proportional to $N_{ij}^{-1}$ and, under some simplifying assumptions, 
 the best way to aggregate noisy observations with unequal noise 
 variance can be seen to be to use the inverse variances as weights in a least-squares criterion.
Also, while  \cite{dalvi_aggregating_2013} focuses on the spectral properties of the task-worker assignment 
matrix (or graph), we focus on the worker-worker assignment graph and its properties.

In a way our approach to skill estimation is to consider a weighted low-rank decomposition, 
and thus our approach falls into the category of moment-based spectral methods.
A spectral approach based on a task-task correlation matrix 
(whose entries give the number of workers that labeled two tasks in the
same manner) is considered by \cite{ghosh2011moderates},
while the belief-propagation algorithm 
of
\cite{karger_iterative_2011} can also be viewed as a power-iteration method 
applied to the task-worker matrix, as the authors themselves note in this paper.
The theoretical results in these last two papers concern random worker-task assignments.
\cite{Zhang2014} propose to use a spectral method to initialize an expectation-maximization 
algorithm (originally due to \cite{DaSke79}) and they provide 
error bounds for both the skills and labels estimated and
for non-random worker-task assignments.
In a recent work \cite{BoCo16} consider a simple method to estimate skills considering triangles of workers
using a moment method. This can be viewed as a simple heuristic approach to approximately solve 
our weighted rank-one approximation problem where the weights are simply disregarded.
All the works mentioned except that of 
\cite{dalvi_aggregating_2013} assume a full worker-task assignment matrix.

\vspace{-0.05in}
\section{Conclusions}
We propose a method for skill estimation for the single-coin crowd-sourcing binary classification model in symmetric noise. 
We define the (worker) interaction graph whose nodes are workers and an edge between two nodes indicates 
whether or not the two workers participated in a common task. 
We show that skills are asymptotically identifiable if and only if an appropriate limiting version of the interaction graph 
is irreducible and has odd-cycles. 
We then formulate a weighted rank-one optimization problem to estimate skills based on observations on the interaction graph. 
We propose a gradient descent scheme, and show that asymptotically 
it converges to the global minimum. We characterize robustness to noise in terms of spectral properties of the interaction graph. We then demonstrate that a plug-in estimator based on the estimated skills achieves state-of-art performance on a number of real-world datasets.

\bibliographystyle{plain}
\bibliography{refs}

\appendix

\newcommand{\x}{{\bf x}}
\newcommand{\p}{{\bf p}}
\newcommand{\y}{{\bf y}}
\newcommand{\bpi}{{\boldsymbol{\pi}}}

\def\N{\mathcal{N}}
\def\ox{\overline{x}}
\def\ovr{\overline{\rho}}
\def\oA{\overline{A}}
\def\A{\mathcal{A}}
\def\B{\mathcal{B}}
\def\oa{\overline{a}}
\def\oG{\overline{G}}
\def\sjn{\sum_{j=1}^n}
\def\E{\mathcal{E}}
\def\oE{\mathcal{\overline{E}}}
\def\oN{\mathcal{\overline{N}}}
\def\0{{\bf 0}}
\def\1{{\bf 1}}
\def\e{{\bf e}}
\def\s{{\bf s}}
\def\R{\mathbb{R}}
\def\C{\mathbb{C}}
\providecommand{\comjh[1]}{\com{JH}{#1}}
\providecommand{\com[2]}{\begin{tt}[#1: #2]\end{tt}}
\def\A{{\mathcal A}}
\def\jsr{{\rm jsr}}
\def\ao#1{\color{red}{#1}}
\def\aoc#1{{\color{red}#1}}

\def\bang#1{\smallbreak\noindent$\triangleright$\ {\it #1}\ }
\def\bangg#1{\smallbreak\noindent$\unrhd$\ \textit{#1}\ \ }

\def\A{\mathcal{A}}
\def\B{\mathcal{B}}
\def\red#1{{\color{red}#1}}
\def\ao{}
\def\aa{}

\def\argmin{\mathop{\rm argmin}}
\def\mf{\mathbf}
\def\mb{\mathbb}
\def\mc{\mathcal}
\def\beq{\begin{equation*}}
\def\eeq{\end{equation*}}
\def\bql{\begin{equation}}
\def\eql{\end{equation}}
\def\bqn{\begin{eqnarray*}}
\def\eqn{\end{eqnarray*}}
\def\bnl{\begin{eqnarray}}
\def\enl{\end{eqnarray}}
\def\bma{\begin{bmatrix}}
\def\ema{\end{bmatrix}}
\def\bmx{\begin{matrix}}
\def\emx{\end{matrix}}
\def\ben{\begin{enumerate}}
\def\een{\end{enumerate}}
\def\bit{\begin{itemize}}
\def\eit{\end{itemize}}
\def\bei{\begin{itemize}}
\def\eei{\end{itemize}}
\def\bet{\begin{tabular}}
\def\eet{\end{tabular}}
\def\und{\underline}
\def\unb{\underbrace}
\def\Log{\mbox{Log}}
\newcommand{\allcaps}[1]{\uppercase\expandafter{#1}}
\newcommand{\yn}{\color{blue}}
\def\cb #1{{\color{blue} #1}}
\def\ao #1{{\color{red} #1}}

\linespread{1.05}         



\section{Proof of Theorem~\ref{thm:learnability}}
The proof directly follows from Lemma~\ref{lem:asylearn} and Lemma~\ref{lem:asylearn1}.
We will next prove these Lemmas.

\noindent
{\it Proof of Lemma~\ref{lem:asylearn}:}
Take any two workers $i,j$ that are connected in $G = ([W],E)$. Let $t\in \N$ be such that $(i,t),(j,t)\in A$.
By assumption, $Y_{i,t} Y_{j,t} = g_t^2 Z_{i,t} Z_{j,t} = Z_{i,t} Z_{j,t}$. 
Now, by the law of large numbers, 
\begin{align*}
C_{ij} \doteq& \lim_{T\to\infty} \frac1T \sum_{(i,t),(j,t)\in A, t\le T}  Z_{i,t} Z_{j,t} = \EE{ Z_{i} Z_j }\\
 =& \EE{ Z_i } \EE{ Z_j } = s_i s_j,
\end{align*}
where $(Z_i)_i \sim \Pi_{i=1}^W \Rad(s_i)$. Note that $C_{ij} = C_{ji}$. We define $C_{ij} = 0$ when $(i,j)\not\in E$.

We note in passing that 
the above system of equations can be written compactly as
\[
M \circ s s^\top = M \circ C\,,
\]
where $M\in \{0,1\}^{W\times W}$ denotes the adjacency matrix of $G$,
$C\in [-1,1]^{W\times W}$ is the matrix formed of $(C_{ij})$ and $\circ$ denotes the entrywise (a.k.a. Hadamard, or Schur) product of matrices.

Now, WLOG assume that workers $1,2,\dots,2k+1$ form a cycle in $G$:  $(1,2),\dots,(2k,2k+1),(2k+1,1)\in E$.
Then, 
\begin{align*}
s_1  &= C_{1,2k+1} s_{2k+1}^{-1} \\
       & = C_{1,2k+1} C_{2k+1,2k}^{-1} s_{2k} \\
	   & = C_{1,2k+1} C_{2k+1,2k}^{-1} C_{2k,2k-1} s_{2k-1}^{-1} \\
	   & \quad  \vdots \\
        &  =  C_{1,2k+1} C_{2k+1,2k}^{-1} C_{2k,2k-1} \dots C_{2,1} s_{1}^{-1} \,,
\end{align*}
or 
\begin{align*}
|s_1| = \sqrt{C_{1,2k+1} C_{2k+1,2k}^{-1} C_{2k,2k-1} \dots C_{2,1}} \,,
\end{align*}
assuming that $C_{2,3},C_{4,5},\dots,C_{2k,2k+1}\ne 0$. 
Since $G$ is connected, for any worker $i$ there exists a path from worker $1$ to worker $i$.
If this path was given by the vertices $1,2,\dots,\ell$ then
\begin{align*}
|s_\ell| =& |C_{\ell,\ell-1}| \,|s_{\ell-1}^{-1} | = |C_{\ell,\ell-1}| \,|C_{\ell-1,\ell-2}^{-1}|\, |s_{\ell-2} |\\
= &\dots = |C_{\ell,\ell-1}| \,|C_{\ell-1,\ell-2}^{-1}| \cdots |C_{2,1}^{(-1)^{\ell}}|\, |s_1|^{(-1)^{\ell+1}}\,.
\end{align*}

It remains to show that $\mathcal{P}(s)$ can be recovered.
Let $i,j\in [W]$ be different workers. Then, if $\pi \subset E$ is a path in $G$ from $i$ to $j$,
we have $\Pi_{(u,v)\in E} \sgn(C_{u,v}) = \Pi_{(u,v)\in E} \sgn(s_u) \sgn(s_v) = \sgn(s_i) \sgn(s_j)$ 
regardless of how $\pi$ is chosen.
Now, if $i$ and $j$ are such that for some path $\pi$ connecting them  $\Pi_{(u,v)\in E} \sgn(C_{u,v}) =+1$,
we assign $i,j$ to the same group. Since $G$ is connected, this creates at most two groups and the resulting
``partition'' must match $\mathcal{P}(s)$. 

\noindent
{\it Proof of Lemma~\ref{lem:asylearn1}:}
We show this by construction.
Without loss of generality, assume that one of the even-cycles is formed by edges$(1,2),\ldots,(2k-1,2k)$. Let us denote $s'_1=2s_1, s'_{2}=\frac{s_2}{2},s'_3=2s_3,s'_4=\frac{s_4}{2},\ldots,s'_{2k-1}=2s_{2k-1},s'_{2k}=\frac{s_{2k}}{2}$. We can verify that
\begin{equation*}
C_{1,2}=s_1s_2=s'_1s'_2,C_{2,3}=s_2s_3=s'_2s'_3,\ldots,C_{2k-1,2k}=s_{2k-1}s_{2k}=s'_{2k-1}s'_{2k}.
\end{equation*}



Note that learnability now follows directly from \cref{lem:asylearn}.

\noindent{\it Reverse Direction:} For the other direction, if $s,s'\in [-1,1]^W$ are different skill vectors such that $|s| = |s'|$ and $\mathcal{P}(s) = \mathcal{P}(s')$ and $s,s'\in S(\Theta')$. It follows that $s=-s'$.
Take any $g\in \pms^W$. Note that the instances $(s,A,g)$ and $(-s,A,-g)$ lead to the same joint distribution over the observed labels. Hence, no algorithm can tell these instances apart, thus any algorithm will suffer linear regret on one of these instances. For this reason , we only consider a subset of learnable instance set $\Theta$ where $\sum_{i\in[W]}s_{i}>0$ in the rest of this paper.

\section{Proof of Theorem~\ref{thm:noiselessPGD}}

\noindent {\bf The problem:} we are given a matrix $N$ which is nonnegative, irreducible, aperiodic, with integer entries, symmetric, and with zero diagonal; and also a vector $s \in \R^W$. Does there exist a vector $x \neq s$ such that, for each $i=1, \ldots, W$, we have
\begin{equation} \label{maineq} \sum_{j=1}^W N_{ij} (x_i x_j - s_i s_j) x_j = 0. \end{equation}

\bigskip

\noindent {\bf 0.} Of course, $x_i=0$ for all $i$ is always a solution, but presumably we are looking for a nonzero solution.

\medskip

\noindent {\bf 1.} Let us adopt the following notation. For a vector $x$, $D_x$ will refer to the diagonal matrix with $x$ on the diagonal. For a matrix $A$, ${\rm diag}\left[ A\right]$ will refer to the diagonal of $A$ stacked as a vector. Also, let us refer to the set of matrices which are nonnegative, irreducible, aperiodic, symmetric entries and with zero diagonal as {\em admissible}.

\smallskip

\noindent {\bf 1.1.} Our  first observation is that we may rewrite Eq. (\ref{maineq}) as
\begin{equation} \label{rewritten} {\rm diag}\left[ N D_x (x x^T - s s^T) \right] = 0. \end{equation}

\medskip

\noindent {\bf 2.} Let us now make the simplifying assumptions that $s > 0$ and that we are looking for $x > 0$. We will later lift this restriction to establish the general result. For this case We will argue that the answer is negative. Specifically, we will argue that given $s > 0$ we cannot find $x > 0, x \neq s$ and admissible $F$ such that
\[ {\rm diag}\left[ F (x x^T - s s^T ) \right]  = 0. \] We were able to drop the $D_x$ from the equation because $N$ is admissible if and only if $N D_x$ is. 

\smallskip

\noindent {\bf 2.1.}  Since 
\[ x_i x_j - s_i s_j = s_i \left( \frac{x_i}{s_i} \frac{x_j}{s_j} - 1 \right) s_j \] defining $u_i = x_i/s_i$ we have that $u>0$ and that
\[ x x^T - s s^T = D_{s} (u u^T - 1 1^T) D_s \] We must therefore argue that it is impossible to find $u>0, u \neq 1$ and admissible $W$ such that 
\[ {\rm diag} \left[ F D_s (u u^T - 1 1^T) D_s \right] = 0\] Since $s>0$ it will suffice to argue that we cannot find $u>0, u \neq 1$ and admissible $Z$ such that 
\[ {\rm diag} \left[ Z (u u^T -1 1^T) \right] = 0.\] 

\noindent {\bf 2.2.} We now complete the proof as follows. Without loss of generality, we can assume that  $u_1 \leq u_2 \leq \cdots \leq u_W$; we can always relabel indices to make this hold. 

Now there are three possibilities:
\begin{enumerate}\item $u_1 u_W > 1$.
\item $u_1 u_W = 1$. 
\item $u_1 u_W < 1$. 
\end{enumerate}

Let us  consider the first possibility. In that case the last column of $u u^T - 1 1^T$ is strictly positive, and therefore, considering $[Z(uu^T - 1 1^T)]_{WW}$, we obtain that the last row of $N$ must be zero --  contradicting irreducibility.  Similarly, in case $3$, the first column of $u u^T - 1 1^T$ is negative, and, considering $[Z(uu^T - 1 1^T)]_{11}$, we see that the first row of $N$ must be zero, which can't be. 

It remains to consider case $2$.  Consider any $u>0, u \neq 1$.  We may assume that $u_1  < u_n$ (ruling out the possibility that $u$ is proportional to the all-ones vector can easily be done separately).

 First, we break up $\{1,\ldots, W\}$ into three blocks. The first block is all the indices $j$ such that $u_j = u_1$. The third block is all the indices $j$ such that that $u_j = u_W$. All the other indices go into block $2$. Note that block 2 may be empty, for example if every entry of $u$ is equal to $u_1$ or $u_W$.

The advantage of partitioning this way is that 
the matrix $u u^T - 1 1^T$ has the following sign structure:

\[ u u^T - 1 1^T = \left( \begin{array}{c|c|c} 
- & - & 0 \\
\hline
- & * & + \\
\hline
0 & + & + 
\end{array} \right)\] where $-$ represents a strictly negative submatrix, $+$ represents a strictly positive submatrix, while $*$ represents a submatrix that can have elements of any sign.  The strict negativity comes from the fact that $u_1 < u_n$. 

 Partitioning $Z$ in the natural way, we have that 
\[ {\rm diag} \left[   \left( \begin{array}{c|c|c} 
Z_{11} & Z_{12} & Z_{13} \\
\hline
Z_{21} & Z_{22} & Z_{23} \\
\hline
Z_{31} & Z_{32} & Z_{33} 
\end{array} \right) \left( \begin{array}{c|c|c} 
- & - & 0 \\
\hline
- & * & + \\
\hline
0 & + & + 
\end{array} \right) \right] = 0.\]

Considering the $(1,1)$ diagonal block of the above product we obtain $Z_{11} = Z_{12}=0$; and considering the $(3,3)$ diagonal block of the above product we obtain  $Z_{32} =  Z_{33}=0$. 

But from here we can easily derive a contradiction. Indeed,  if the second block is nonempty, the matrix is reducible; and if the second block is empty, it is periodic with period two. 

We are now left to prove that there are no additional critical points. Suppose $\sgn(x_is_i)$ is neither all positive or negative. Our goal is to arrive at a contradiction. 
We again consider the matrix formed by $Q_{ij}(x,s)=\sum_j N_{ij} (x_ix_j-s_is_j)x_j=0$. In matrix form we can write it as:
$$ 
Q(x,s)=[Q_{ij}(x,s)]=(\text{diag}[x_i/s_i] N \text{diag}[x_i/s_i] – N) [\text{vec}([x_is_i])] = 0
$$ 
where $\text{vec}([x_is_i])]$ denotes a vector whose components are $x_is_i$.  
Without loss of generality let the first j components be negative and the rest positive. Note that $x_i/s_i$ and $x_is_i$ have the same sign. So we now examine the last W-j rows of the matrix $Q(x,s)$ and notice that they align with the sign pattern of $\text{vec}([x_is_i])]$. Consequently, unless the matrix is reducible this product must be positive and we arrive at a contradiction. Consequently $\sgn(x_is_i)$ must all be the same. Our earlier analysis for the case $s_i >0, x_i>0$ now applies. Indeed, we note that since $x_is_i$ and $u=x_i/s_i$ must have the same sign, without loss of generality we can assume this to be positive. Now as in {\bf 2.1} we let $F=ND_xD_s$ and $u_i$ as before and note that $F$ is still admissible. The proof then follows as long as no component of $s_i$ is zero. Therefore, we have either that $x_i=s_i$ or $x_i=-s_i$ are the only equilibrium points if no component of $s$ is zero. 

 While we used our identity weighting to establish our result a close examination reveals that the proof generalizes to any non-negative weights as well. This proves the first part of the theorem. The asymptotic case is now a simple consequence of the law of large numbers and the usual limiting arguments.

\section{Proof of Theorem~\ref{thm:noisyPGD}}


We deal with the case of $x>0$ and $s>0$. The other cases follow in a similar manner. For future reference and notational simplicity we denote by $s_{\min}=\min_i |s_i|$ and $\sigma_{\min} \triangleq \sigma_{\min}(P(1))$. 
\noindent {\bf Definition:}  For a vector $x$, $D_x$ will refer to the diagonal matrix with $x$ on the diagonal. For a matrix $A$, ${\rm diag}\left[ A\right]$ will refer to the diagonal of $A$ stacked as a vector. Also, let us refer to the set of matrices which are nonnegative, irreducible, not bipartite, symmetric, with integer entries and with zero diagonal as {\em admissible}.

\bigskip

\noindent {\bf Definition:} Let $G=(V,E)$ be an undirected  graph and let $p: E \rightarrow \R_{+}$ be a positive weight function which is symmetric: $p(\{i,j\})=p(\{j,i\})$. The signless Laplacian is defined to be  the matrix $L$ whose entries are
\[ L_{ij} = p_{\{i,j\}}, ~~~ L_{ii} = \sum_{j=1}^W p_{\{i,j\}}. \] It is well known that the signless Laplacian is positive semidefinite, and positive definite if and only if $G$ is not bipartite \cite{desai94}.

\bigskip

\noindent {\bf Theorem A:} Let $N$ to be some admissible matrix as described in Theorem~\ref{thm:noisyPGD}. For each $\epsilon \in (0,1)$ there exists a constant $c_{\epsilon} > 0$ with the following property: if $\Delta \in \R^{W \times W}, x \in \R^W$ satisfy
\begin{eqnarray} {\rm diag}\{ND_x (x x^T - s s^T + \Delta)\} & = & 0 \label{maineq} \\ 
x_i & \in & [\epsilon, 1/\epsilon] ~~~~~~~ i = 1, \ldots, n \nonumber \\
\|\Delta\|_2 & \leq & c_{\epsilon} \nonumber
\end{eqnarray} then 
\begin{equation} \label{concl} \|x - s\|_2 \leq c \|\Delta\|_2 \end{equation} where 
\[ c= 2 \frac{\|N\|_{\rm Fro}}{\mu} \frac{1}{(s_{\min})^2}, \] where $\mu$ is the smallest eigenvalue of the signless Laplacian of the graph with weights $N_{ij}$. 

%

\bigskip

Before proceeding to the proof, we need several preliminary lemmas.

\bigskip

\noindent {\bf Lemma B:} (local strong convexity of the objective) Set \begin{equation} \label{fdef} f(x) = \sum_{i<j} N_{ij} (x_i x_j - s_i s_j)^2, \end{equation} where $s>0$.
There is some $\delta >0$ such that  \[ \lambda_{\rm min} \left( \nabla^2 f(x) \right) \geq  \mu (\min_i s_i)^2  \] whenever $x$ and $s$ are positive vectors satisfying 
\[ \frac{x_i}{s_i} \in [1-\delta, 1+\delta],  ~~~ i = 1, \ldots, n. \] Here $\mu$ is the same as in Theorem A, the smallest eigenvalue of the signless Laplacian.

\bigskip

\noindent {\bf Proof of Lemma B:} Indeed, 
\[ \frac{\partial f}{\partial x_i} = \sum_{j=1}^W 2 N_{ij} (x_i x_j - s_i s_j) x_j, \] and
\begin{eqnarray*} \frac{\partial^2 f}{\partial x_i^2} & = & \sum_{j=1}^W 2 N_{ij} x_j^2 \\
\frac{ \partial^2 f}{\partial x_i x_j} & = & 4 N_{ij} x_i x_j - 2 s_i s_j N_{ij}, ~~~~~ i \neq j.
\end{eqnarray*}

For a vector $u \in \R^W$, define $P(u)$ to be the matrix 
\begin{eqnarray*} P_{ii}(u) & = & \sum_{j=1}^W 2 N_{ij} u_i^2 \\ 
P_{ij(u)} & = & 4 N_{ij} u_i u_j - 2 N_{ij}
\end{eqnarray*} Then 
\[ \nabla^2 f(x) = D_s P(x./s) D_s, \] so \[ \lambda_{\rm min} \left( \nabla^2 f(x) \right) \geq (\min_i s_i)^2 \lambda_{\rm min}\left( P(x./s) \right). \] This implies that 
\[ \lambda_{\rm min} \left(  \nabla^2 f(s) \right) \geq (\min_i s_i)^2 \lambda_{\rm min} (P(1)), \]  and the last quantity is exactly $2\mu$. The lemma now follows  because $\lambda_{\rm min}(P(u))$ is a continuous function of $u$.

\bigskip

\noindent {\bf Lemma C:} Fix an admissible $N$ and positive vector $s$.  Define 
\[ U_{\epsilon, r} = \{ x ~|~ \|x./s - 1\|_{\infty} \geq r, ~~~ x_i \in [\epsilon, 1/\epsilon], ~~ i = 1, \ldots, n\} \]
and   
\[ c_{\epsilon,r} =  \inf_{\{  x \in U_{\epsilon, r}\}} ~ \|{\rm diag}(N D_x (x x^T - s s^t))\|_{2} \] Then \[ c_{\epsilon, r} > 0. \]

\bigskip

\noindent {\bf Proof of Lemma C:} Immediate by continuity. 

\bigskip

\noindent {\bf Proof of Theorem A:}

Suppose that assumptions of the theorem hold. For a given $\epsilon$, choose the quantity  $c_{\epsilon}$ small enough so that \[ \|N\|_{\rm Fro} \frac{1}{\epsilon} c_{\epsilon}  < c_{\epsilon, \delta'}, \] where (i) $\delta' =\min(\delta, 1)$ and $\delta$ was defined in Lemma $B$ (ii) $c_{\epsilon, \delta}$ was defined in Lemma C.

Now since the assumptions of the theorem hold, we have 
\[ {\rm diag} \left( N D_x (x x^T - s s^T) \right) = {\rm diag} (N D_x \Delta) \] 
\begin{equation} \label{twosides} \| {\rm diag}\{N D_x (x x^T - s s^T)\}\|_{2} \leq \|N\|_{\rm Fro} (\max_i x_i) \|\Delta\|_2  \leq \|N\|_{\rm Fro} \frac{1}{\epsilon} c_{\epsilon}  < c_{\epsilon, \delta'} \end{equation}

By definition of $c_{\epsilon,\delta'}$ this lets us conclude that $x./s \notin U_{\epsilon, \delta'}$. Since, by assumption, $x_i \in [\epsilon, 1/\epsilon]$ for all $i$, we have that  $\|x./s - 1\|_{\infty} \leq \delta'$. 

Since Lemma B established that the $f(x)$ from Eq. (\ref{fdef}) is strongly convex in the set $\{x ~|~ \|x./s - 1\|_{\infty} \leq \delta\}$, and since we have just established that $x$ belongs to this set (and $s$ belongs to it automatically), we can use the following well-known inequality: 
\[ (\nabla f(x) - \nabla f(s))^T (x - s) \geq \mu (\min_i s_i)^2\|x - s\|_2^2. \] Using the fact that $\nabla f(s) = 0$ and applying Cauchy-Schwarz, we obtain 
\[ \|\nabla f(x)\|_2 \|x - s\|_2  \geq \mu (\min_i s_i)^2 \|x-s\|_2^2 \] or 
\begin{eqnarray*} \|x-s\|_2 & \leq & \frac{1}{\mu (\min_i s_i)^2} \|\nabla f(x)\|_2 \\ 
& = & \frac{1}{\mu (\min_i s_i)^2}  \|{\rm diag}(N D_x (x x^T - s s^T)) \|_2 \\ 
& = & \frac{1}{\mu (\min_i s_i)^2}  \|{\rm diag} (N D_x \Delta) \|_2 \\ 
& \leq & \frac{2}{\mu} \|N\|_{\rm Fro} \frac{\max_i s_i}{(\min_i s_i)^2} \|\Delta\|_2,
\end{eqnarray*} where in the last step  we used $\delta' \leq 1$ along with the bound $$\max_i x_i \leq (1+\delta') \max_i s_i \leq 2 \max_i s_i.$$

While we used our identity weighting to establish our result a close examination reveals that the proof generalizes to any non-negative weights as well. This proves the first part of the theorem. 

\section{Additional Experiments}
\subsection{Experiments for Different Skill-Distribution}
We randomly assign binary classes to $T=300$ tasks and select five pairs of parameters. Average prediction errors are presented in Table~\ref{Tab:beta} averaged over $10$ independent runs. Parameters $\alpha=5,\beta=1$, correspond to reliable workers leading to small prediction error; the prediction error with parameters $\alpha=2,\beta=2$ and $\alpha=0.5,\alpha=0.5$, is almost random because of $\sum_{i\in [W]}s_{i}$ is no longer positive, which validates our theory. Similar situation arises for $\alpha=2,\beta=5$ and $\alpha=5,\beta=1$, because the skills are all flipped relative to our assumption that the sum of the skills is positive. 

\begin{table}
\caption{Average prediction errors with different skills distributions.}
\label{Tab:beta}
\vskip 0.15in
\centering
\scalebox{0.85}{
\begin{tabular}{c|c|c|c|c}
\hline
Type of workers                & $\alpha$  & $\beta$   & Bayes error           & Prediction error (const. noise) \\  \hline
Adversary vs. hammer            & $0.5 $    & $0.5$     &$ 0.0036\pm 0.0014$    &$ 0.5990\pm 0.4860$ \\ \hline
Asym. with more positive skills & $5 $      & $1$       &$ 0.0038\pm 0.0014$    &$ 0.0041\pm 0.0013$ \\ \hline
Asym. with more negative skills & $2 $      & $5$       &$ 0.0314\pm 0.0062$    &$ 0.9667\pm 0.0067$ \\ \hline
Hammer                          & $2 $      & $2$       &$ 0.0615\pm 0.0083$    &$ 0.4162\pm 0.4273$ \\ \hline
Spammer                         & $1 $      & $3$       &$ 0.0129\pm 0.0034$    &$ 0.9864\pm 0.0041$ \\ \hline
\end{tabular}}
\end{table}

\subsection{Graph Size}
We focus on how the graph size affects the performance of PGD algorithm. Note that graph size is associated with the number of workers. Our goal is to demonstrate that for a constant amount of noise, prediction accuracy of PGD does not degrade with graph-size. We again consider the case when the worker-interaction graph is a star-graph with an odd-cycle of length 3. We increase the size of worker-interaction graph by adding nodes to the star-graph. Skills $s$ are selected between $0.8$ and $-0.3$ uniformly. 
To fix the noise level, we define $C_{ij}=s_is_j+\xi_{ij},\forall (i,j)\in E$ where $\xi_{ij}$ is randomly selected from $[-0.2,0.2]$. Note that the noise level is quite large relative to what we expect in terms of accuracy of correlation estimates. We iteratively run PGD for $50$ times. The average prediction errors with different graph size it presented in Table~\ref{Tab:graphsize}. It can be seen that the prediction error is not sensitive to the graph size compared to the Bayes error.
\begin{table}
\caption{Average prediction errors for different graph sizes.}
\label{Tab:graphsize}
\vskip -0.1in
\centering
\scalebox{0.85}{
\begin{tabular}{c|c|c|c|c}
\hline
  Number of workers & $21$  & $51$ & $71$ & $91$  \\ \hline
Bayes error & $0.0425\pm{0.0042}$  & $0.0622\pm{0.0040}$ & $0.0634\pm{0.0033}$ & $0.0574\pm{0.0030}$ \\ \hline
Prediction error (const. noise)& $0.0425\pm{0.0042}$ & $0.0641\pm{0.0126}$ & $0.0662\pm{0.0072}$ & $0.0618\pm{0.0063}$\\ \hline
\end{tabular}}
\end{table}

\end{document}